# Unsupervised Language-agnostic WER Standardization

*Satarupa Guha\*, Rahul Ambavat\*, Ankur Gupta, Manish Gupta, Rupeshkumar Mehta*

Microsoft

{saguha, raambava, angup, gmanish, rupeshme}@microsoft.com

*Equal contribution

## Abstract

Word error rate (WER) is a standard metric for the evaluation of Automated Speech Recognition (ASR) systems. However, WER fails to provide a fair evaluation of human perceived quality in presence of spelling variations, abbreviations, or compound words arising out of agglutination. Multiple spelling variations might be acceptable based on locale/geography, alternative abbreviations, borrowed words, and transliteration of code-mixed words from a foreign language to the target language script. Similarly, in case of agglutination, often times the agglutinated, as well as the split forms, are acceptable. Previous work handled this problem by using manually identified normalization pairs and applying them to both the transcription and the hypothesis before computing WER. In this paper, we propose an automatic WER normalization system consisting of two modules: spelling normalization and segmentation normalization. The proposed system is unsupervised and language agnostic, and therefore scalable. Experiments with ASR on 35K utterances across four languages yielded an average WER reduction of 13.28%. Human judgements of these automatically identified normalization pairs show that our WER-normalized evaluation is highly consistent with the perceived quality of ASR output.

## 1. Introduction

This Word error rate (WER) is a common metric of the performance of an automated speech recognition (ASR) system. It is computed by first aligning the recognized word sequence with the reference (spoken) word sequence using dynamic string alignment. WER is then the ratio of total number of word insertion, deletions or substitutions to number of words in the reference. WER works at a word level and assumes that there is *one* standard way of transcribing every word in the target language.

However, there are many scenarios where the assumption does not hold: (1) ASR on code-mixed speech samples. (2) Spell variations based on locale/geography. (3) Widely accepted abbreviations or shortened forms. (4) Evolution of spellings over time. (5) Inconsistency in spellings of borrowed words from other languages. (6) Agglutinated versus split forms of compound words. We discuss these scenarios in detail in the following.

Most spoken conversations in multi-lingual communities are highly code-mixed. Transliteration of foreign words is based on phonetics, and therefore changes with varying pronunciations/dialects/accents. Since there is no standard way of spelling foreign words, there are many variations of the same word in the target language. E.g., consider this Hindi sentence written in Devanagari script: कुत्ता एक डोमेस्टिक जानवर है. Here the word 'domestic' can be spelled as डॉमेस्टिक or as डोमेस्टिक. Table1 shows examples of other such spell variations.

Table 1: *Examples of spelling variations because of transliteration from English to another language*

| Original Word | Valid Variations | Target Language |
|---|---|---|
| Domestic | डोमेस्टिक; डॉमेस्टिक | Hindi |
| Insurance | इन्शुरन्स; इंस्युरेन्स | Marathi |
| Duty | ડયૂટિ; ડયૂટી | Gujarati |
| Home | ప్లాం; ప్లాం | Telugu |

Annotated transcriptions could be noisy especially for low resource languages (LRL) due to typing limitations in modern keyboards or software used by human experts [1]. Spell variations could also be based on locale/geography, e.g., American vs British spelling forms (rationalise/rationalize, color/colour), variants of Hindi (बढ़िया/बढ़ियाँ). Temporal spelling variations are also common, e.g. गाँव/गांव are both acceptable; the latter has grown in prominence recently.

Often, words from one language make their way into another language's vocabulary; we call such words as borrowed words. In many such cases, these words contain sounds uncommon in target language. To support the written forms of these borrowed words, two approaches are generally taken: (a) map to the character representing the closest sound, or (b) introduce a new character by modifying an existing character whose sound is similar. For example, the word zubaani (जुबानी) is borrowed into Hindi from Urdu. Since the z sound is not originally present in Hindi, it is sometimes denoted by the closest character ज (j) and spelt as जुबानी (jubaani), while others use the diacritic mark (nukta) in conjunction with ज to denote this same sound by ज़ and spell it as ज़ुबानी.

Widely accepted abbreviations or shortened words also lead to spell variations, e.g., catalogue/catalog, programme/program. Another source of variation is in display forms of compound words, which are formed by joining two or more in-vocabulary words. This problem is especially severe in languages with a high degree of agglutination. Many times, both the compound form as well as separate words (bigrams, trigrams, etc.) are acceptable, e.g., in Marathi, आईवडील (mother father) vs आई वडील (mother father); in English, both 'speakerphones' as well as 'speaker phones' are both popular. Also, in some languages, it is valid to split a word with morphological postpositions (suffixes) into two separate words, e.g., in Marathi: संदीपला, संदीपचे, vs संदीप ला and संदीप चे, where ला and चे are the postpositions. Similarly,

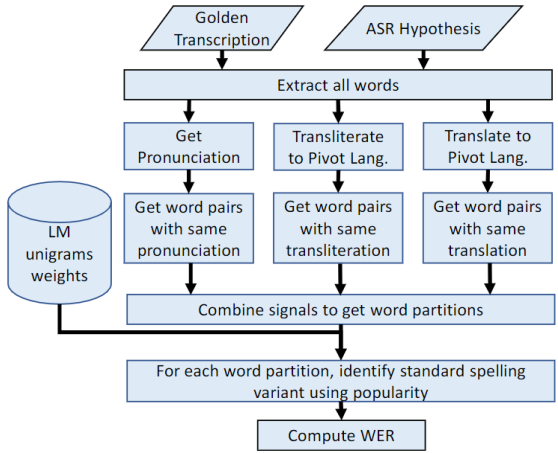

Figure 1: *Spelling Normalization technique*

in Gujarati, અમદાવાદનો vs અમદાવાદ નો and in Telugu, తెలుగులో vs తెలుగు లో are both valid.

Due to these variations, there are often a significant number of differences between a golden transcription and ASR output pair, leading to WER that is misleadingly high. It is very difficult to evaluate an ASR system based on such a misleading metric especially in case of languages with many such variations. Spurious system comparisons can lead to wasted debugging/development effort, deployment of systems with inferior perceived quality, etc.

Debugging and OOV analysis becomes complicated, since mismatch between transcription and hypothesis might influence a developer to add the mismatched words from the transcription into the vocabulary, but in reality, an acceptable variation of the same word might already be present in the vocabulary. Manually identifying such words is very cumbersome and not a good use of developer's time.

In this paper, we make the following main contributions: (1) We propose a system for standardization of transcriptions and ASR output aimed at computing a more reliable WER. (2) Our solution consists of two main modules: spelling normalization and segmentation normalization as illustrated in Figs. 1 and 2 respectively. The proposed system is unsupervised, language agnostic and scalable. (3) Our experiments with four languages lead to an average WER reduction of 13.28% along with high consistency between normalized WER using our system and human perceived quality.

## 2. Spelling normalization

We show the flow of our spelling normalization module in Fig. 1. Given a reference and a hypothesis from an ASR system, we apply spelling normalization at word level on both, converting them to a more standard form. The first step to achieve this is to identify word partitions or equivalence classes such that all words within a partition are interchangeable. One way to obtain equivalent words is to align the reference and the hypothesis instances from a large dataset, and then consider aligned word pairs as candidates. But alignment can be noisy and can propagate errors to the downstream steps. Hence, we consider all words from both the transcriptions and the hypotheses from the entire corpus as initial potential candidates.

In absence of any labeled data for training a model that would identify spelling variations, we derive weak supervision via proxy related tasks. We hypothesize that for a pair of words that are variations, ideally, they should have the same (1) pronunciation, (2) transliteration to another language, and (3) translation to another language.

We hypothesize that for a pair of words that are equivalent, ideally, they should have the same pronunciation, e.g., 'colour' and 'color' both same the same pronunciation ('k ah l ah r'), ट्वेंटी and ट्वेन्टी both are pronounced as 'tr vw ay n tr iy'. In some cases, specifically in the context of code-mixed utterances, the pronunciation of equivalent foreign words varies slightly. Table 2 highlights some examples of acceptable pronunciation variations. To overcome this, we apply the relaxation of certain phone pairs, similar to [2]. One caveat of this approach is that it cannot distinguish between homophones, since by definition, they have the same pronunciation but are different from a semantic point of view, e.g., करता (meaning: 'does') and कर्ता (meaning: 'doer') have the same pronunciation ('k ah r t a').

We hold similar hypothesis for transliteration as well, i.e., equivalent word pairs typically have the same transliteration, e.g., टुमारो and टुमॉरो both result in the same transliterated English word 'tomorrow'. While effective, this technique cannot be applied in isolation since there exist word pairs specially in code-mixed scenarios which are semantically different but have the same transliteration. E.g., Hindi word मैं (meaning: 'I') when transliterated to English cannot be differentiated from the English word 'main' (written as मेन in Devanagari). Similarly, ही (Hindi word meaning 'only') and हाई (meaning 'hi') both have the same English transliteration 'hi'.

In the same vein, the translations of equivalent words should ideally be the same. Evidence can be found in the following examples: स्टुडियो and स्टुडिओ are both rightly translated to the English word 'studio', आउटलूक and आऊटलुक both map to the English word 'outlook' as expected. Notably, the homophone pair करता and कर्ता are translated to different English words 'does' and 'doer' respectively, making them distinguishable, which both pronunciation and transliteration alone failed to. On the other hand, there are cases where translation cannot handle alone, in absence of the other two signals. E.g., the Hindi word अर्थ and the English word मीनिंग both are translated to the English word 'meaning', although they are entirely distinct words.

For pronunciation, we use an in-house grapheme-to-phoneme (G2P) system trained on a large amount of data, typically used as a component of the hybrid ASR setup. For each word, we obtain a phone sequence from this system. We use the publicly available Microsoft Azure Cognitive Services Translator API (https://azure.microsoft.com/en-in/services/cognitive-services/translator/) for both translation and transliteration. In all cases, we use English as the intermediate pivot language. Our experimental results show that a combination of signals from these three systems when aggregated lead to an impressive 99% precision for the task of identifying valid spelling variants.

Table 2: *Pronunciation variations for equivalent words.*

| Word 1 | Pron 1 | Word 2 | Pron 2 | Meaning |
|---|---|---|---|---|
| डॉमेस्टिक | dr oh m ay s tr ih k | डोमेस्टिक | dr ow m ay s tr ih k | Domestic |
| मल्टीपल | m ah l tr iy p ah l | मल्टिपल | m ah l tr ih p ah l | Multiple |
| कम्प्यूटर | k ah m p y uw tr ah r | कम्प्युटर | k ah m p y uh tr ah r | Computer |

Once the spelling variants are obtained, we use unigram weights in the language model (LM) for the target language to determine the most popular display form. We replace all other variants by the one with the highest LM weight. Since in most cases, all the spelling variations are acceptable, this step is optional., all the spelling variations are acceptable, this step is optional.

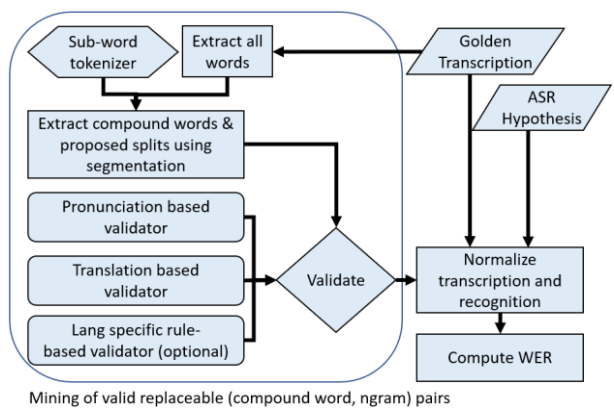

Figure 2: *Segmentation normalization techniques*

## 3. Segmentation normalization

We show the flow of our segmentation normalization module in Fig. 2. Similar to the spelling normalization approach, we apply the segmentation normalization at the word level for both the transcription and the ASR output. A naïve way to handle postposition (suffix) based compound words is to create a whitelist of possible suffixes for that language and join all those words from the suffix list with base words [3]. However, this approach is not scalable as there are 150+ possible suffixes in Marathi, for example, and also this requires language expertise. Additionally, this cannot handle non-suffix compound words.

In order to alleviate these issues, as shown in Fig. 2, we propose the following solution: (1) Extract all distinct words from transcription. (2) For each word, segment, validate (using three different validation methods), and gather a list of valid (compound word, segmented ngram) pairs. (3) Replace all segmented ngrams in transcription and recognition by their corresponding compound words from this list.

Segmentation or tokenization is a common task in NLP that breaks down a piece of text into smaller units called tokens, e.g., 'handbag' → 'hand' + 'bag'. There are various tokenization algorithms available that can be used for our task such as, Morfessor segmentation [4], SentencePiece [5], Byte-pair encoding [6], WordPiece [7], etc. We found that results using any of these tokenization methods do not differ much, and hence report ones for Morfessor only, for lack of space.

While tokenization algorithms split words as appropriate based on frequency of those splits in the training corpus, not all splits are valid words by themselves and hence are not meaningful for our use case. Hence, we apply a set of validations to make sure that we only consider meaningful splits: (1) Pronunciation-aware segmentation, (2) Translation-aware segmentation, and (3) (optionally) Language specific rule-based validation.

Pronunciation-aware segmentation involves ensuring that pronunciation of the compound word is the same as pronunciation of the corresponding ngram. For example, tokenizer splits 'subscription' as {'subscript', 'ion'}. However, this is correctly considered to be an invalid split for our use case, since the pronunciation of 'subscription' is not same as the combination of pronunciations of 'subscript' and 'ion'. On the other hand, splitting of 'Nehruji' by the tokenizer into {'Nehru', 'ji'} is rightly considered valid based on the same criterion.

Translation-aware segmentation involves ensuring that translation of the compound word to a pivot language is the same as translation of the corresponding ngram to the pivot language. We used English as our pivot language. For example, tokenizer splits ऑलवेज़ (translation: always) as {ऑल, वेज़} (translation: {'all', 'ways'}), and this is rightly considered invalid.

Finally, a few language specific validation tests can be optionally employed. One example is a rule which specifies that no token can begin with a diacritic (a glyph added to a letter.).

## 4. Experiments and results

We tested our module on the two most dominant language families in the Indian subcontinent - Indo-Aryan (IA) and Dravidian (D). Experiments were run on the following languages: Hindi (IA), Marathi (IA), Gujarati (IA) and Telugu (D). Hindi, Marathi, Gujarati and Telugu are four popular Indian languages with a combined speaker base of around 850M people. For each language, we test on a proprietary phrasal set created from real user data for the purpose of evaluation of our in-house ASR systems. The dataset includes short speech samples, 4.82 seconds on average. There are a total of 12002 utterances for Hindi, 7339 for Marathi, 6164 for Gujarati and 8838 for Telugu. We report aggregated WER per language. While the proposed technique is applicable to any ASR setup, our experiments were based on a hybrid system consisting of an HMM-LCBLSTM (Hidden Markov Model with latency-controlled BLSTM) based acoustic model and a 5-gram LM with backoff smoothing, both trained on proprietary data. We trained a Morfessor segmentation model for each locale with a vocabulary size of 1M words.

Table 3: *WER Reduction (WERR) by Spelling and Segmentation Normalization.*

| Lang | Utt | Base WER | Spell Norm | | Seg Norm | |
|---|---|---|---|---|---|---|
| | | | WER | WERR | WER | WERR |
| Hin | 12002 | 12.1 | 10.5 | 12.8 | 10.6 | 11.8 |
| Mar | 7339 | 18.2 | 17.9 | 1.8 | 15.7 | 13.6 |
| Guj | 6164 | 19.3 | 18.6 | 3.6 | 18.8 | 2.6 |
| Tel | 8838 | 32.5 | 28.8 | 11.2 | 30.6 | 5.7 |

In Table 3, for each language, we show base WER (WER without our normalization methods), Norm WER, as well as WER reduction WERR = (Norm WER-Base WER) / Base WER. Across languages and test sets, we observe a reduction in WER ranging from ~2% to ~13% owing to spelling normalization, and that ranging from ~3% to ~14% owing to segmentation normalization. We also experimented with a cascading approach where we first performed segmentation normalization and then spelling normalization for both transcription as well as ASR recognition. That leads to even better WERR values of 13.71, 15.58, 6.23 and 17.53 for Hindi, Marathi, Gujarati and Telugu respectively.

Table 4: *Manual Judgement of Spelling Normalization and Segmentation Normalization pairs.*

| Language | Spelling Norm | | Segmentation Norm | |
|---|---|---|---|---|
| | Pairs | Accuracy | Pairs | Accuracy |
| Hindi | 280 | 100% | 76 | 100% |
| Marathi | 191 | 99.95% | 88 | 100% |
| Gujarati | 421 | 99.56% | 136 | 99.30% |
| Telugu | 622 | 98.87% | 396 | 99.7% |

The spelling and segmentation normalization pairs obtained by our system were reviewed by language experts for correctness - to verify if the extracted pairs can be used interchangeably irrespective of context. For example, in the mined pair {"Nehru ji","Nehruji"}, the language expert is tasked to judge if "Nehru ji" can be replaced by "Nehruji" in all cases. Based on their review, we observed an accuracy of ~99% across all languages which shows that the pairs we mined are high quality, and normalized WER reported with these pairs are more accurate. Accuracy here would be the fraction of times judges marked the mined pairs as acceptable out of all the mined pairs.

As shown in Table 4, our proposed system leads to a WER computation which is highly correlated with human perceived quality since those variations in hypothesis (with respect to transcription) would not be considered as errors by humans.

## 5. Related work

### 5.1. Spelling Correction and Checking

Previous work in literature has mainly focused on spell correction or spell check. Spell correction models have been trained using labeled data with (correct, incorrect) term pairs [8]. Such data is scarce for LRLs, and hence there is hardly any work on spell correction for LRLs. Etoori et al. [9] resorted to synthetic labeled data for training their models. Spell checking on the other hand is a relatively simpler task and has been attempted for LRLs as well [10]. However, in this paper, rather than spell correction or spell checking, we focus on the problem of spell normalization where multiple spell variations are acceptable.

### 5.2. Spell Normalization

In the context of code-mixed ASR where mixed script output often leads to high WER, transliteration optimized WER metric has been introduced in [11]. Our method handles not just transliteration-based spell variants, but also cares about other variants based on pronunciation and translation. Singh et al. [12] propose a normalization technique based on word embeddings which requires per-language manual tuning of a similarity threshold for Levenshtein distance.

### 5.3. Segmentation Normalization

In order to avoid segmentation normalization on ASR output, some papers ([13], [14]) resort to handling it in LM part of ASR itself using sub-word based LMs. Unlike them, we propose a more generic system which handles segmentation normalization along with spell normalization on top of any ASR output as well for golden transcriptions, thereby providing a more holistic solution for WER normalization.

## 6. Conclusions

WER can lead to misleading evaluation of ASR systems in languages with many acceptable word variations. In this work, we presented an extensive list of causes for such variations. To incorporate all such causes in a realistic WER evaluation, we proposed a system with two important modules (spelling normalization and segmentation normalization) which in turn leverage state-of-the-art NLP tools like tokenization, pronunciation, transliteration and translation. Experiments on four different languages show an average WER reduction of 13.28%. Our proposed system is unsupervised and language agnostic, and thus easy to scale to more languages. Application to two widely differing language families (in terms of their origin as well as syntactic and grammatical constructs) show that the approach is generic. In the future, we would like to investigate the effect of this normalization on the training data and in turn, its impact on LM quality.